\documentclass{article}
\usepackage[a4paper, total={5.5in, 8.3in}]{geometry}
\usepackage[hyphens]{url}                           
\usepackage[svgnames]{xcolor}                       
\usepackage{hyperref}                               
\usepackage[T1]{fontenc}                            
\usepackage[utf8]{inputenc}                         
\usepackage{amsmath}                                
\usepackage{mathrsfs}                               
\usepackage{amsthm}                                 
\usepackage{amssymb}                                
\usepackage{amsfonts}                               
\usepackage{leftidx}                                
\usepackage{float}                                  
\usepackage{graphicx}                               
\usepackage[labelfont=bf]{caption}
\usepackage{subcaption}
\usepackage{array}                                  
\usepackage{makecell}                               
\usepackage[noend, ruled, onelanguage]{algorithm2e} 
\newcolumntype{P}[1]{>{\centering\arraybackslash}p{#1}} 

\newcommand{\pa}[1]{\left(#1\right)}
\newcommand{\pac}[1]{\left[#1\right]}

\let\OLDthebibliography\thebibliography
\renewcommand\thebibliography[1]{
  \OLDthebibliography{#1}
  \setlength{\parskip}{0pt}
  \setlength{\itemsep}{0pt plus 0.3ex}
}

\title{\line(1,0){400}\\ \textbf{\textsc{SignSGD}: Fault-Tolerance to Blind and Byzantine Adversaries}\\\line(1,0){400}}
\author{
  \textbf{Jason Akoun, Sébastien Meyer} \\
  École Polytechnique, France \\
  \texttt{\{firstname.lastname\}@polytechnique.edu}
}
\date{}

\begin{document}

\maketitle

\begin{abstract}
  Distributed learning has become a necessity for training ever-growing models by sharing calculation among several devices. However, some of the devices can be faulty, deliberately or not, preventing the proper convergence. As a matter of fact, the baseline distributed SGD algorithm does not converge in the presence of one Byzantine adversary. In this article we focus on the more robust \textsc{SignSGD} algorithm derived from SGD. We provide an upper bound for the convergence rate of \textsc{SignSGD} proving that this new version is robust to Byzantine adversaries. We implemented \textsc{SignSGD} along with Byzantine strategies attempting to crush the learning process. Therefore, we provide empirical observations from our experiments to support our theory.  Our code is available on GitHub\footnote{\url{https://github.com/jasonakoun/signsgd-fault-tolerance}} and our experiments are reproducible by using the provided parameters.
\end{abstract}

\section{Introdution}

With the increasing size of datasets and of the diversity of their sources, the need for large-scale distributed systems has never been so important. In the field of distributed learning, there are two types of distributed settings. The first setting is centralized, that is, a server gathers gradients computed locally on the devices and broadcasts back the changes to make to local models. The second one is decentralized, with the information about model parameters having to propagate from device to device. Moreover, the learning process can happen synchronously or asynchronously. Typical examples of distributed centralized settings are the supercomputers that train state-of-the-art deep learning models. Decentralized asynchronous settings usually happen with small and abundant devices that are not switched on at the same time, such as phones. In the case of phones, there are also models that are centralized but fine-tuned locally (think about your phone's auto-completion of words). In this project, we focused on the centralized synchronous setting. Among the "workers" or "processes", there can be adversaries. All types of adversaries are included in this denomination, from the unintentional faulty processes to the coordinated, omniscient adversaries. 

\vspace{\topsep} 

The main issue of the learning task is to avoid the propagation of faults onto the workers. Indeed, the classical stochastic gradient descent algorithm is not fault-tolerant, as we will show later on. Therefore, a gradient descent algorithm must provide the same dynamic of convergence as in Bottou 1998\cite{bottou1998}, that is, the aggregated gradient must fall in the decreasing half-space of the loss function. One of the most successful gradient descent algorithm that as been proposed in the recent years is Krum, and was detailed in 2017 by Blanchard et al.\cite{krum2017}. Nevertheless, the proposed algorithm and theoretical bounds for the convergence rate only work for a proportion of Byzantine adversaries bounded by $\mathcal{O}(\sqrt{d})$ where $d$ is the space dimension and the learning process is more difficult with non convex loss functions. Despite the fact that there has been several follow-ups to this paper, other alternatives have been developed. In their 2018 paper, Bernstein et al.\cite{signsgd-optimisation} have proposed a new gradient descent algorithm, namely \textsc{SignSGD}. In 2019, Bernstein et al.\cite{signsgd-fault-tolerant} extended \textsc{SignSGD} to \textsc{Signum} and proved the theoretical tolerance of both algorithms to blind adversaries. 

\vspace{\topsep} 

In this article, we recall the most important results from the initial papers and we try to go further by proposing a more general theoretical bound for the convergence rate of \textsc{SignSGD}, as well as experimental results to support our claims.

\section{Previous work}

In this section, we mainly recall results and propositions from both the initial paper\cite{signsgd-optimisation} and the extension to fault-tolerance\cite{signsgd-fault-tolerant}. When looking at a particular algorithm for gradient descent, we want to verify the following properties:

\begin{itemize}
  \item[] \textbf{D1.} Fast algorithmic convergence
  \item[] \textbf{D2.} Good generalisation performance
  \item[] \textbf{D3.} Communication efficiency
  \item[] \textbf{D4.} Robustness to network faults  
\end{itemize}

Clearly, it will be unreasonable to think that one can devise an algorithm satisfying all four properties with high certainty. The usual stochastic gradient descent algorithm does satisfy the \textbf{D1} and \textbf{D2} properties, and this explains why it has been so widely used in machine and deep learning. Regarding \textbf{D3}, the stochastic gradient descent algorithm needs to communicate full vectors of gradients from workers to servers and the other way around. In addition, \textbf{D4} is not verified for several cases. Consider the example of an omniscient adversary. This adversary would just have to send to the server the inverse sum of the gradients values of all the other processes in order to stop the training. Thus, the authors have proposed a new algorithm, namely \textsc{Signum}, based on the communication of gradients signs. 

\vspace{\topsep} 

\begin{algorithm}[H]
  \SetAlgoLined
  \LinesNumbered
  \KwIn{learning rate $\eta > 0$, momentum $\beta \in [0, 1)$, weight decay $\lambda \geq 0$, batch size $n$, initial point $x$, number of workers $M$.}

  Initialize momentum $v_m \leftarrow 0$ for each worker\;
  \Repeat{convergence (or criterion)}
  {
    \ForEach{worker $m$}
    {
      $\widetilde{g}_m \leftarrow \frac{1}{n} \sum \limits_{i=1}^n F_i(x)$\;
      $v_m \leftarrow (1 - \beta)\widetilde{g}_m + \beta v_m$\;
      \textbf{push} $\text{sg}(v_m)$ \textbf{to} server\;
    }
    \For{the server}
    {
      $V \leftarrow \sum \limits_{m=1}^M \text{sg}(v_m)$\;
      \textbf{push} $\text{sg}(V)$ \textbf{to} workers\;
    }
    \ForEach{worker $m$}
    {
      $x \leftarrow x - \eta (\text{sg}(V) + \lambda x)$\;
    }
  }
  \caption{\textsc{Signum} with majority vote. All operations are element-wise. Setting $\beta = 0$ yields \textsc{SignSGD}.}
\end{algorithm}

\vspace{\topsep} 

It appears that the proposed algorithm verifies \textbf{D3} by communicating only signs between devices. Also, the \textbf{D2} property stems naturally from this simple algorithm. We will now look at both \textbf{D1} and \textbf{D4} properties.

\subsection{Assumptions}

The authors proved in their paper a theoretical bound for the convergence rate of \textsc{SignSGD}. They use four assumptions, of which the first three are usual assumptions in papers concerning gradient descent algorithms.

\vspace{\topsep} 

\noindent \textbf{Assumption 1.} (Lower bound) \textit{For all $x$ and some constant $f^*$, we have objective value $f(x) \geq f^*$.}

\vspace{\topsep} 

\noindent \textbf{Assumption 2.} ($L$-Smooth) \textit{Let $g(x)$ denote the gradient of the objective $f(.)$ evaluated at point $x$. Then, $\forall x, y$ we require that for some non-negative constant $L = (L_1, ..., L_d)$,}

\begin{center}
  $$| f(y) - [f(x) + ^{t}g(x)(y-x)] | \leq \frac{1}{2} \sum \limits_i L_i(y_i - x_i)^2$$
\end{center}

\vspace{\topsep} 

\noindent \textbf{Assumption 3.} (Variance bound) \textit{Upon receiving query $x \in \mathbb{R}^d$, the stochastic gradient oracle gives us an independent, unbiased estimate $\widetilde{g}$ that has coordinate bounded variance:}

\begin{center}
  $$\mathbb{E}(\widetilde{g}(x)) = g(x) \quad \mathbb{E}((\widetilde{g}(x)_i - g(x)_i)^2) \leq \sigma_i^2$$
\end{center}

\noindent \textit{for a vector of non-negative constants $\sigma = (\sigma_1, ..., \sigma_d)$.}

\begin{figure}
  \centering
  \captionsetup{type=figure}
  \includegraphics[width=0.7\linewidth]{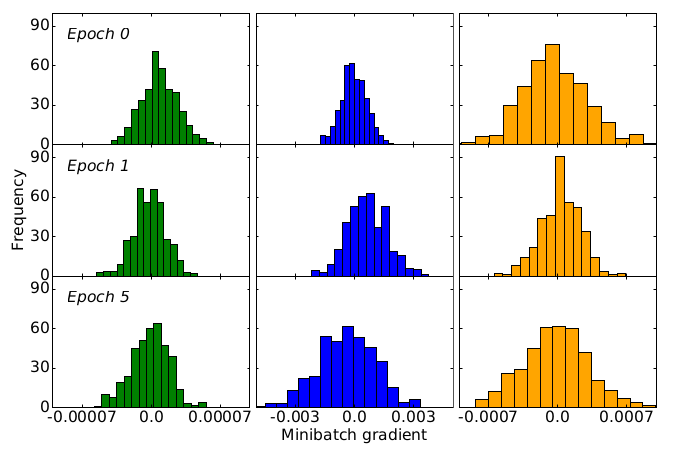}
  \captionof{figure}{Gradients distributions for ResNet18 on CIFAR-10\cite{signsgd-fault-tolerant}.}
  \label{fig:graddist}
\end{figure}

The fourth assumption is less common. The authors assume that the gradients follow unimodal gaussian distributions. This assumption stems from empirical observations, as shown \textbf{Figure \ref{fig:graddist}}.

\vspace{\topsep} 

\noindent \textbf{Assumption 4.} (Unimodal, symmetric gradient noise) \textit{At any given point $x$, each component of the stochastic gradient vector $\widetilde{g}(x)$ has a unimodal distribution that is also symmetric about the mean.}

\subsection{Blind adversaries}

In their original paper, the authors have considered blind adversaries, that is, adversaries that do not know about the gradients of other workers. Since \textsc{SignSGD} algorithm relies on the communication of gradients signs, all the strategies that a blind adversary can think of come down to the following definition.

\vspace{\topsep} 

\noindent \textbf{Definition 1.} (Blind adversaries) \textit{A blind adversary may invert their stochastic gradient estimate $\widetilde{g}_t$ at iteration $t$.}

The first result which allows the authors for proving their upper bound on convergence rate relies on \textbf{Assumptions 3 and 4}.

\vspace{\topsep} 

\noindent \textbf{Lemma 1.} (Bernstein et al., 2018\cite{signsgd-optimisation}) \textit{Let $\widetilde{g}_i$ be an unbiased stochastic approximation to gradient component $g_i$, with variance bounded by $\sigma_i^2$. Further assume that the noise distribution is unimodal and symmetric. Define signal-to-noise ratio $S_i = \frac{|g_i|}{\sigma_i}$. Then we have that} 

\begin{center}
  $$
  \mathbb{P}(\text{sg}(\widetilde{g}_i) \neq \text{sg}(g_i)) \leq 
  \begin{cases}
    \frac{2}{9} \frac{1}{S_i^2} \quad \textit{if} \ S_i > \frac{2}{\sqrt{3}}, \\
    \frac{1}{2} - \frac{S_i}{2 \sqrt{3}} \quad \textit{otherwise}
  \end{cases}
  $$
\end{center}

\noindent \textit{which is in all case less than or equal to $\frac{1}{2}$.}

\vspace{\topsep} 

The bound gives an estimation of the ability to estimate a good approximation of the gradient component knowing that there is a certain noise. It allows to estimate an upper bound for the convergence rate of \textsc{SignSGD}.

\vspace{\topsep} 

\noindent \textbf{Theorem 2.} (Non-convex convergence rate of majority vote with adversarial workers, Bernstein et al., 2019\cite{signsgd-fault-tolerant}) \textit{Run \textbf{Algorithm 1} for $K$ iterations under \textbf{Assumptions 1 to 4}. Switch off momentum and weight decay ($\beta = \lambda = 0$). Set the learning rate, $\eta$, and mini-batch size, $n$, for each worker as} 

\begin{center}
  $$
  \eta = \sqrt{\frac{f_0 - f^*}{||L||_1 K}}, \qquad n = K.
  $$
\end{center}

\noindent \textit{Assume that a fraction $\alpha < \frac{1}{2}$ of the $M$ workers behave adversarially according to $\textbf{Definition 1}$. Then majority vote converges at rate:}

\begin{center}
  $$
  \pac{\frac{1}{K} \sum \limits_{k=0}^{K-1} \mathbb{E}(||g_k||_1)}^2 \leq \frac{4}{\sqrt{N}} \pac{\frac{1}{1 - 2\alpha} \frac{||\sigma||_1}{\sqrt{M}} + \sqrt{||L||_1 (f_0 - f^*)}}^2
  $$
\end{center}

\noindent \textit{where $N = K^2$ is the total number of stochastic gradient calls per worker up to step $K$.}

\vspace{\topsep} 

For further proofs and materials, we link the interested reader to \cite{signsgd-optimisation} and \cite{signsgd-fault-tolerant}.

\section{Extension to Byzantine adversaries}

The previous lemma and theorem that we presented are designed to answer to the question of tolerance to blind adversaries. A more general type of adversaries are the Byzantine adversaries.

\vspace{\topsep} 

\noindent \textbf{Definition 2.} (Byzantine adversaries) \textit{A Byzantine adversary may send an arbitrary value to the server. It is aware of the gradients values of the other workers and it may collude with other Byzantine adversaries to set up a strategy.}

\vspace{\topsep} 

A more general definition of Byzantine adversaries as well as the concept of $(\alpha, f)$-Byzantine resilience can be found in Blanchard et al.\cite{krum2017}. Clearly, Byzantine adversaries are much more dangerous than blind adversaries. In the case of basic stochastic gradient descent, a Byzantine adversary can send a gradient of infinite norm and therefore crush the learning process. In this section, we propose a new upper bound for the tolerance of \textsc{SignSGD} to any type of adversaries. Moreover, we will only make use of \textbf{Assumptions 1 to 3}.

\vspace{\topsep} 

\noindent \textbf{Lemma 1bis.} \textit{Let $\widetilde{g}_i$ be an unbiased stochastic approximation to gradient component $g_i$, with variance bounded by $\sigma_i^2$. Define signal-to-noise ratio $S_i = \frac{|g_i|}{\sigma_i}$. Then, we have that}

\begin{center}
  $$
  \mathbb{P}(\text{sg}(\widetilde{g}_i) \neq \text{sg}(g_i)) \leq \frac{1}{2 S_i^2}
  $$
\end{center}

\begin{proof}
  It is a direct application of Bienaymé-Tchebychev's inequality.
\end{proof}

With this new lemma, we are able to prove a new bound for the convergence rate of \textsc{SignSGD}.

\vspace{\topsep} 

\noindent \textbf{Theorem 2bis.} \textit{Run \textbf{Algorithm 1} for $K$ iterations under \textbf{Assumptions 1 to 3}. Switch off momentum and weight decay ($\beta = \lambda= 0$). Set the learning rate, $\eta$, and mini-batch size, $n$, for each worker as}

\begin{center}
  $$
  \eta = \sqrt{\frac{f_0 - f^*}{||L||_1 K}}, \qquad n = K.
  $$
\end{center}

\noindent \textit{Assume that a fraction $\alpha < 1 - 1/2p$ of the $M$ workers behave adversarially according to $\textbf{Definition 2}$. Then majority vote converges at rate:}

\begin{center}
  $$
  \pac{\frac{1}{K} \sum \limits_{k=0}^{K-1} \mathbb{E}(||g_k||_1)}^2 \leq \frac{4}{\sqrt{N}} \pac{\frac{1}{2 \sqrt{2}} \frac{1}{p (1-\alpha) - \frac{1}{2}} \frac{||\sigma||_1}{\sqrt{M}} + \sqrt{||L||_1 (f_0 - f^*)}}^2
  $$
\end{center}

\noindent \textit{where $p = \mathbb{P}(\text{sg}(\widetilde{g}_t) = \text{sg}(g_t))$ and $N = K^2$ is the total number of stochastic gradient calls per worker up to step $K$.}

\begin{proof}
  Denote by $M$ the total number of workers, by $\alpha$ the proportion of Byzantine workers, by $Z_t$ the number of correct bits received by the server at iteration $t$ and by $Z_t^g$ the number of bits sent by healthy workers and received by the server at iteration $t$.

  In the worst case, Byzantine adversaries are omniscient and know about the true sign of the gradient. Therefore, they oppose to it. In this case, only healthy workers can help finding the true sign of the gradient. So $\mathbb{P}(Z_t \leq \frac{M}{2}) \leq \mathbb{P}(Z_t^g \leq \frac{M}{2})$.

  Now, $Z_t^g \hookrightarrow \textit{Binomial}((1-\alpha)M, p)$ where $p = \mathbb{P}(\text{sg}(\widetilde{g}_t) = \text{sg}(g_t))$, hence

  \begin{align*}
    \mathbb{P}(Z_t \leq \frac{M}{2}) &\leq \mathbb{P}(Z_t^g \leq \frac{M}{2}) && \text{(Worst case)} \\
    & = \mathbb{P}(\mathbb{E}(Z_t^g) - Z_t^g \geq \mathbb{E}(Z_t^g) - \frac{M}{2}) && \mathbb{E}(Z_t^g) > \frac{M}{2} \\
    & \leq \frac{1}{1 + \frac{\pa{\mathbb{E}(Z_t^g) - \frac{M}{2}}^2}{\text{Var}(Z_t^g)}} && \text{(Cantelli's inequality)} \\
    & \leq \frac{1}{2} \frac{\sqrt{\text{Var}(Z_t^g)}}{\mathbb{E}(Z_t^g) - \frac{M}{2}} && 1 + x^2 \geq 2x \\ 
    & = \frac{1}{2} \frac{\sqrt{p(1 - p)(1 - \alpha)}}{p(1 - \alpha) - \frac{1}{2}} \frac{1}{\sqrt{M}} \\
    & \leq \frac{1}{2} \frac{\sqrt{1-p}}{p(1 - \alpha) - \frac{1}{2}} \frac{1}{\sqrt{M}} && p(1-\alpha) \leq 1 \\
    & \leq \frac{1}{2 \sqrt{2}} \frac{1}{p(1-\alpha) - \frac{1}{2}} \frac{1}{S_i \sqrt{M}} && (\textbf{Lemma 1bis})
  \end{align*}

  The next stage of the proof relies on the same elements as in \cite{signsgd-fault-tolerant}, that is, we compute a telescoping sum over the iterations, and we use our bound to majorize one of the terms.
\end{proof}

\vspace{\topsep} 

\noindent \textbf{Remark 1.} The condition $\mathbb{E}(Z_t^g) > \frac{M}{2}$ can be written as $\alpha < 1 - \frac{1}{2p}$ and implies that $\alpha < \frac{1}{2}$ and $p > \frac{1}{2}$.

\vspace{\topsep} 

\noindent \textbf{Remark 2.} The probability of failure in estimating the true sign of the gradient decreases as the number of workers $M$ increases, when $\alpha$ is fixed.

\vspace{\topsep} 

\noindent \textbf{Remark 3.} If $p = 1$, we do obtain a probability of failure equal to zero. This is coherent with the fact that healthy workers do not make mistakes and are in majority.

\vspace{\topsep} 

Finally, we see that our bound is more general than the one from \textbf{Theorem 2}, however we had to introduce a new parameter $p$. This value measures the ability of estimating the true sign of the gradient and it can depend on many things, such as the dataset.

\section{Implementation}

We implemented a basic distributed SGD as well as \textsc{Signum} in Python. We decided to follow the \textit{PyTorch}\cite{torch} support and we implemented classes for our datasets, optimizers and neural networks with distributed support\cite{torchdist}. Experiments can be run through command lines for logistic and linear regressions with simple feed-forward networks, MNIST\cite{mnist} with two different neural networks and ImageNet\cite{imagenet} with ResNet18 or ResNet50\cite{resnet}. 

\vspace{\topsep} 

Then, we designed a Byzantine strategy for both the distributed SGD and \textsc{Signum} algorithms. In the case of distributed SGD, one Byzantine worker is enough to stop the learning process. This adversary can invert the sum of the gradients of all the other workers and thus eliminate the gradient. In the case of \textsc{Signum}, the Byzantine adversaries will need to collude. First, they collect the gradients signs of all the other workers. Then, they compute the local sum of these signs to estimate if they can beat the healthy workers. Let $f$ be the number of Byzantine adversaries and $s^h$ the sum of gradients signs for healthy workers. For each coordinate $i$, if $s_i^h > f$ or $s_i^h < -f$, the Byzantines cannot invert the final sign, therefore they just oppose to the other workers. If $f >= s_i^h >= 0$, $f - s_i^h$ Byzantine workers will send $-1$, then the other Byzantine adversaries will send $-1$ and $+1$ one after another, starting with $-1$, to try to kill the sign. If $0 > s_i^h >= -f$, they do the same starting with $+1$. Clearly, the resulting learning process will depend on the result of the operation $\text{sg}(0)$. In \textit{PyTorch}, the operation results in $\text{sg}(0) = 0$.

\vspace{\topsep} 

In order to optimize the optimizer steps, we used several tricks. We considered that, amongst the Byzantine adversaries, one is selected to be the Byzantine server and it gathers the gradients signs from the healthy workers. Then, in order to limit the number of communications between processes, the Byzantine server sends the whole Byzantine strategy summed to $f$ while the other Byzantine workers send empty tensors. By doing so and by devising operations on \textit{PyTorch} tensors, the computation time of the optimizer steps with and without Byzantine adversaries are similar. This allows for faster training of the models, as we ran our experiments under CPU.

\section{Experimental results}

The experimental parameters are as follows: $\eta = 10^{-3}$ for distributed SGD and decreases by a factor 10 every 30 steps; $\eta = 10^{-4}$ ($10^{-5}$ for MNIST) for \textsc{SignSGD} and decreases by a factor 10 every 30 steps; $\eta = 10^{-4}$ ($10^{-5}$ for MNIST) and $\beta = 0.9$ for \textsc{Signum} and $\eta$ decreases by a factor 10 every 30 steps. The seed was $8005$ across all experiments. We compared the efficiency of the optimizers on basic datasets which are linear and logistic regressions along with simple feed-forward networks. It is still possible to run experiments on more complex datasets such as MNIST, however they will run on CPU and should take longer. 

\vspace{\topsep} 

\begin{figure}[H]
  \centering
  \captionsetup{type=figure}
  \includegraphics[width=0.85\linewidth]{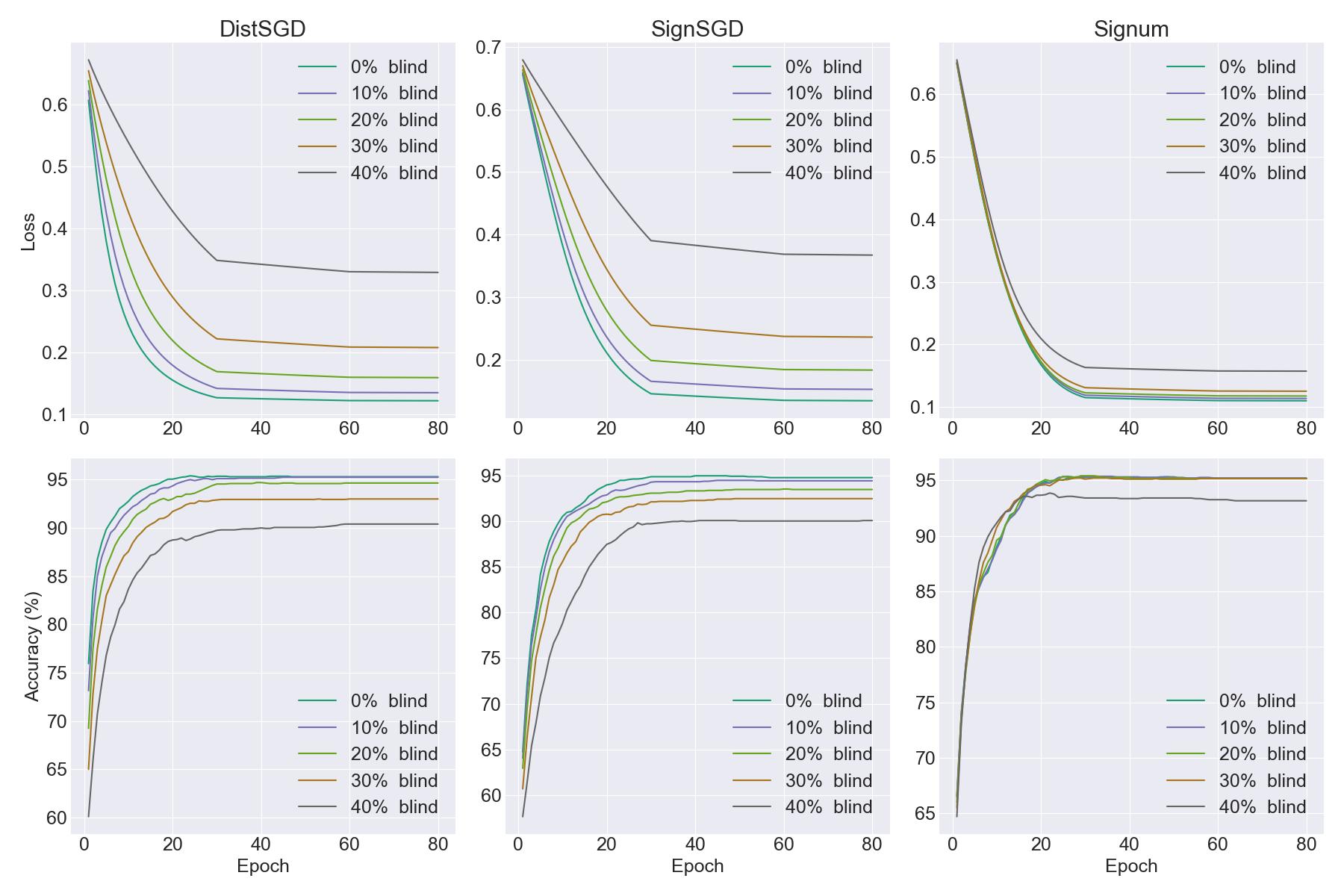}
  \captionof{figure}{Evolution of loss and accuracy for logistic regression with blind adversaries.}
  \label{fig:logregblind}
\end{figure}

Firstly, \textbf{Figure \ref{fig:logregblind}} shows the evolution of accuracy and loss for a logistic regression problem, when there are variable numbers of blind adversaries inverting their gradient signs. From this graph, we can deduce that blind adversaries do not prevent the models from learning. The \textsc{SignSGD} algorithm allows to maintain a better accuracy overall with the number of blind adversaries increasing, and \textsc{Signum} reduces their effect even more. Still, it is important to keep in mind that our dataset and model are basic, therefore the learning process is globally easy.

\vspace{\topsep} 

\begin{figure}[H]
  \centering
  \captionsetup{type=figure}
  \includegraphics[width=0.85\linewidth]{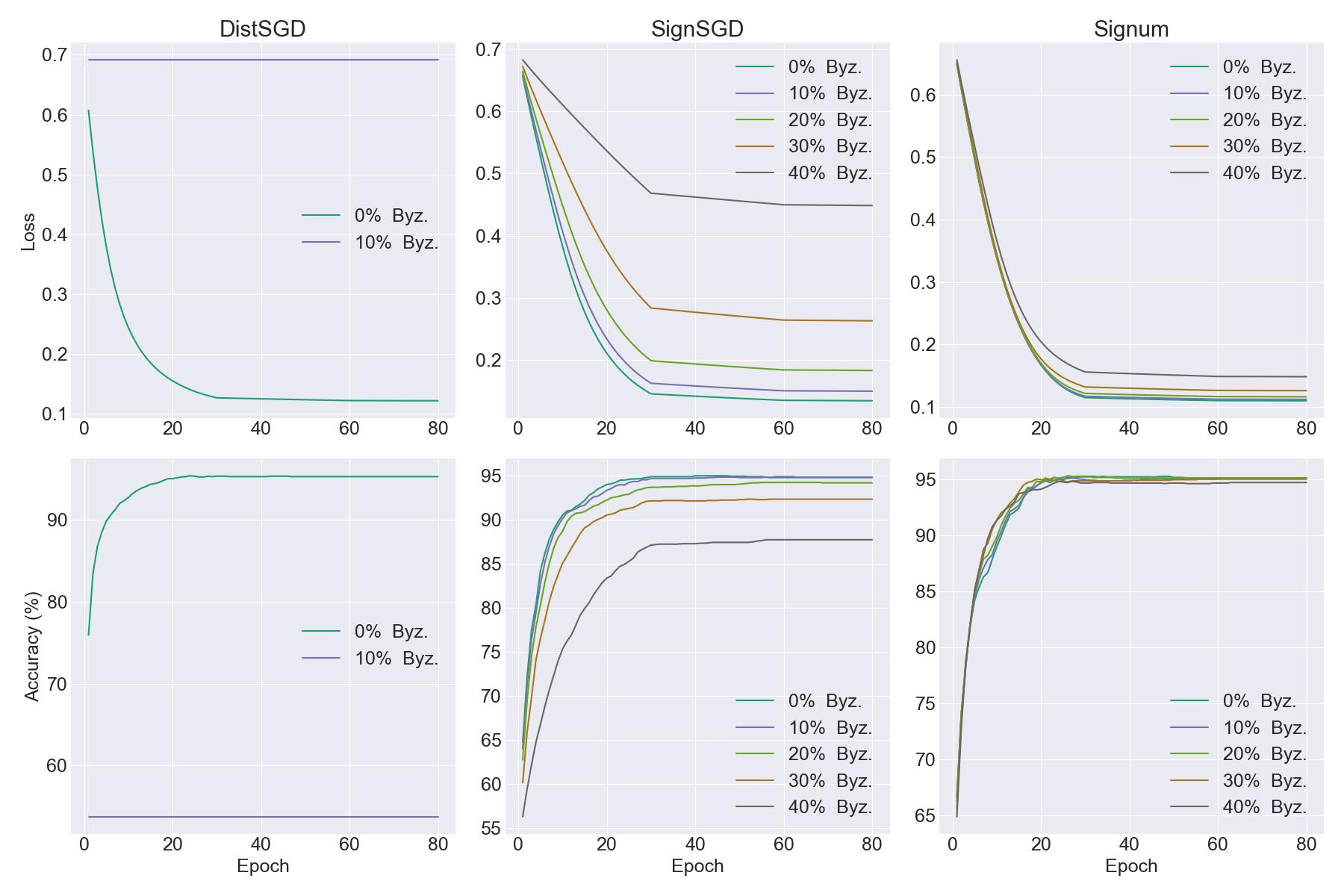}
  \captionof{figure}{Evolution of loss and accuracy for logistic regression with Byzantine adversaries.}
  \label{fig:logregbyz}
\end{figure}

Then, \textbf{Figure \ref{fig:logregbyz}} shows the evolution of loss and accuracy when there are variable numbers of Byzantine adversaries. Byzantine adversaries intercept the gradients of the workers and deploy a strategy. Recall that in the case of distributed SGD, a Byzantine can send arbitrary vectors and thus stop the learning process, and in the case of \textsc{SignSGD}, Byzantine adversaries are limited to sending signs, therefore they try to bring the aggregation to zero. Here, we see that our Byzantine strategy does not break \textsc{SignSGD}. Even more, the \textsc{Signum} version of the algorithm resists to our attacks.

\vspace{\topsep} 

\begin{figure}[H]
  \centering
  \captionsetup{type=figure}
  \includegraphics[width=0.85\linewidth]{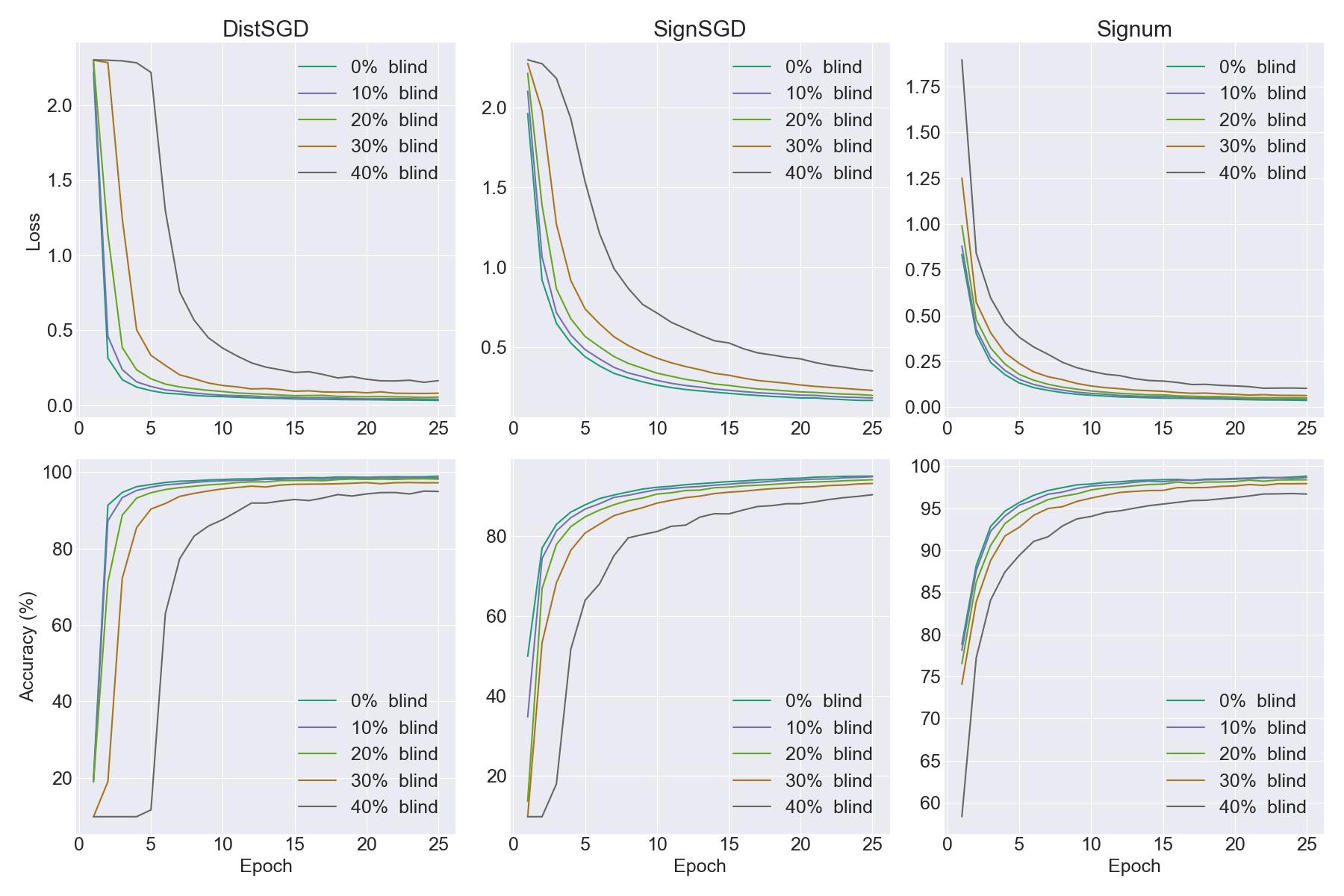}
  \captionof{figure}{Evolution of loss and accuracy for MNIST dataset with blind adversaries.}
  \label{fig:mnistblind}
\end{figure}

The second experiment that we ran was on MNIST dataset. This dataset is much more complex than a logistic regression problem, as it is an image classification task. In the case of blind adversaries, \textbf{Figure \ref{fig:mnistblind}} shows that distributed SGD can resist to the attacks. However with increasing proportion of blind adversaries such as 30\% and 40\%, the learning process takes much more time. SignSGD, and more efficiently Signum, allow to reduce the effect of blind adversaries and to achieve good accuracy, although smaller than the accuracy reached with distributed SGD.

\vspace{\topsep} 

\begin{figure}[H]
  \centering
  \captionsetup{type=figure}
  \includegraphics[width=0.85\linewidth]{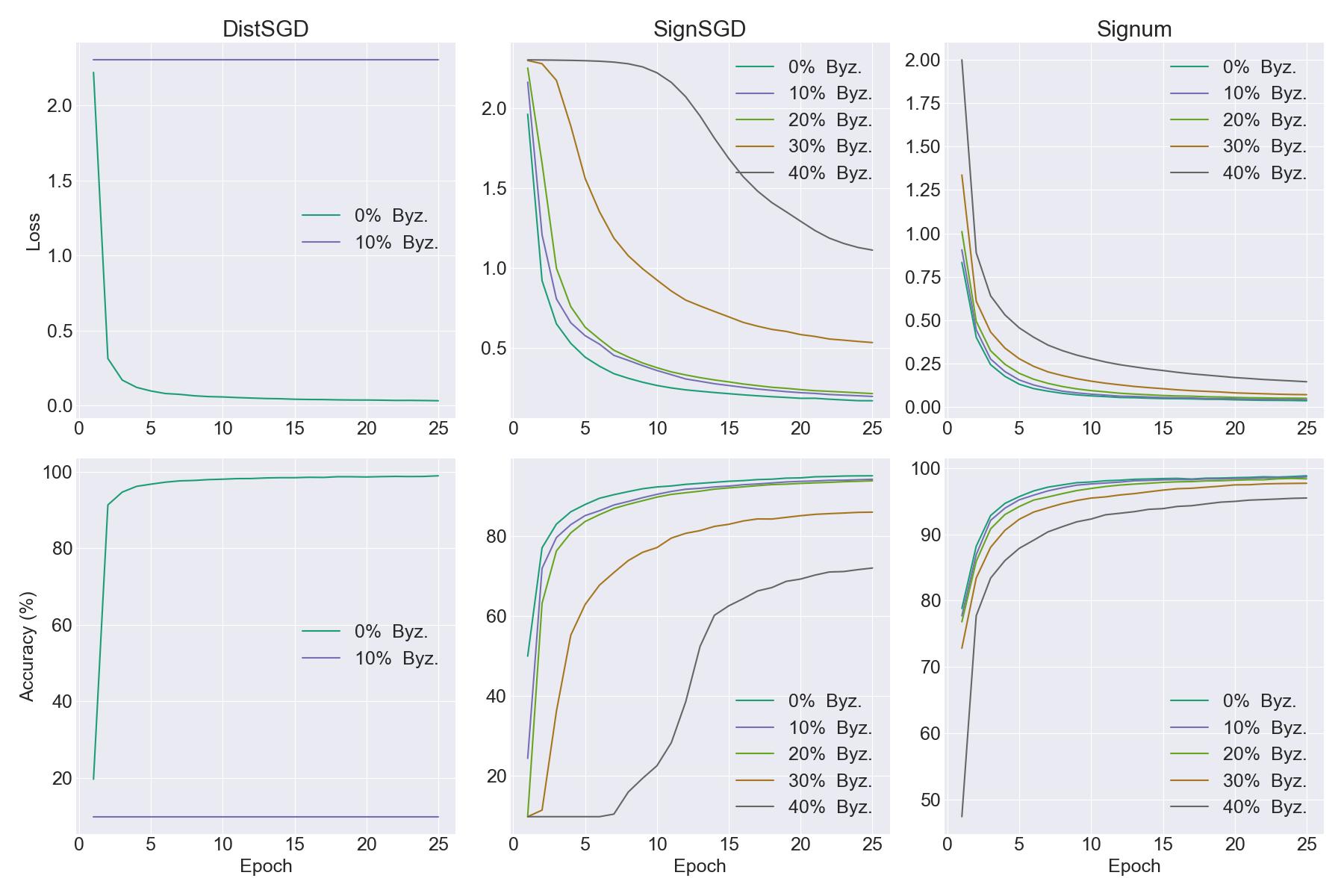}
  \captionof{figure}{Evolution of loss and accuracy for MNIST dataset with Byzantine adversaries.}
  \label{fig:mnistbyz}
\end{figure}

Lastly, \textbf{Figure \ref{fig:mnistbyz}} shows the evolution of loss and accuracy on MNIST when there are variable numbers of Byzantine adversaries. When there are more than 30\% of Byzantine adversaries, it appears that SignSGD is less efficient, however it still allows to learn from the data with decreasing accuracy. Finally, Signum is much more fault-tolerant than SignSGD, as the algorithm allows to achieve an accuracy similar to the one with distributed SGD, even with a proportion of Byzantine adversaries close to 50\%.

\section{Conclusion}

All in all, we have illustrated on simple examples that our new and more general theoretical bound from \textbf{Theorem 2bis} is verified in practice. However, more complex models and data might lead to more difficult situations for the \textsc{Signum} algorithm. Therefore, it might be needed to devise other algorithms to counter specific situations. Furthermore, we have observed that the \textsc{Signum} algorithm implies an overfitting more frequently than other optimizers, since the norm of the aggregation made by the server is not proportional to the loss. 

Further research has been conducted on \textsc{Signum}. We link the interested reader to two other publications on the subject, namely to Jin et al.\cite{stochsgd} where the authors prove a more precise theoretical bound for Byzantine workers than ours when the fourth assumption is not verified, and to Sohn et al.\cite{eleccoding} where the authors devise a new algorithm to protect \textsc{SignSGD} from Byzantine attacks with intermediary servers and prove an associated theoretical bound more precise and asymptotically similar to ours.

\newpage


\begin{thebibliography}{9}

\bibitem{bottou1998}
Léon Bottou. \textit{Online Learning and Stochastic Approximations}. 1998.

\bibitem{krum2017}
Peva Blanchard, El Mahdi El Mhamdi, Rachid Guerraoui and Julien Stainer. \textit{Machine Learning with Adversaries: Byzantine Tolerant Graident Descent}. 2017.

\bibitem{signsgd-optimisation}
Jeremy Bernstein, Yu-Xiang Wang, Kamyar Azizzadenesheli and Anima Anandkumar. \textit{SignSGD: Compressed Optimisation for Non-Convex Problems.} August 2018.

\bibitem{signsgd-fault-tolerant}
Jeremy Bernstein, Jiawei Zhao, Kamyar Azizzadenesheli and Anima Anandkumar. \textit{SignSGD with Majority Vote is Communication Efficient and Fault Tolerant.} February 2019.

\bibitem{torch}
Adam Paszke et al. \textit{PyTorch: An Imperative Style, High-Performance Deep Learning Library}. 2019. Advances in Neural Information Processing Systems, vol. 32, pp. 8024-8035.

\bibitem{torchdist}
Li et al. \textit{PyTorch Distributed: Experiences on Accelerating Data Parallel Training}. 28 June 2020.

\bibitem{mnist}
Li Deng. \textit{The mnist database of handwritten digit images for machine learning research}. 2012. IEEE Signal Processing Magazine, vol. 29, n°6, pp. 141-142.

\bibitem{imagenet}
Jia Deng, Wei Dong, Richard Socher, Li-jia Li, Kai Li and Li Fei-Fei. \textit{Imagenet: A large-scale hierarchical image database}. 2009. IEEE Conference on computer vision and pattern recognition, pp. 248-255.

\bibitem{resnet}
Kaiming He, Xiangyu Zhang, Shaoqing Ren and Jian Sun. \textit{Deep Residual Learning for Image Recognition}. December 2015.

\bibitem{stochsgd}
Richeng Jin, Yufan Huang, Xiaofan He, Huaiyu Dai and Tianfu Wu. \textit{Stochastic-Sign SGD for Federated Learning with Theoretical Guidelines}. September 2021.

\bibitem{eleccoding}
Jy-yong Sohn, Don-Jun Han, Beongjun Choi and Jaekyun Moon. \textit{Election Coding for Distributed Learning: Protecting SignSGD against Byzantine Attacks}. October 2020.

\end{thebibliography}
\end{document}